\documentclass[sn-mathphys]{sn-jnl}% Math and Physical Sciences Reference Style
\usepackage{graphicx}%
\usepackage{multirow}%
\usepackage{amsmath,amssymb,amsfonts}%
\usepackage{amsthm}%
\usepackage{mathrsfs}%
\usepackage[title]{appendix}%
\usepackage{xcolor}%
\usepackage{textcomp}%
\usepackage{manyfoot}%
\usepackage{booktabs}%
\usepackage{algorithm}%
\usepackage{algorithmicx}%
\usepackage{algpseudocode}%
\usepackage{listings}%
\newcommand{\abs}[1]{\left\lvert#1\right\rvert}

\usepackage{caption}

\usepackage{wasysym}
\usepackage{array}
\usepackage{hyperref}

\usepackage[utf8]{inputenc}
\usepackage[T1]{fontenc}

\def\BibTeX{{\rm B\kern-.05em{\sc i\kern-.025em b}\kern-.08em
    T\kern-.1667em\lower.7ex\hbox{E}\kern-.125emX}}

\begin{document}

\title{AI-Based Thesis Assessment: 
An Empirical Study of Human Evaluation Priorities and Their Impact on Automated Assessment}


\author{\fnm{Garv Vikram} \sur{Gursahaney}}%\email{garv-vikram.gursahaney@iu-study.org}

\author{\fnm{Baskhad} \sur{Idrisov}}%\email{baskhad.idrisov@gmail.com}

\author{\fnm{Thorsten} \sur{Fröhlich}}\email{thorsten.froehlich@iu.org}

\author{\fnm{Tim} \sur{Schlippe}}\email{tim.schlippe@iu.org}

\affil{\orgdiv{IU International University of Applied Sciences}, \country{Germany}}


\abstract{
Rubric-based AI systems for thesis assessment use criterion weights to assign different levels of importance to evaluation criteria. These weights are typically defined through expert judgment, although little empirical evidence exists regarding how thesis supervisors actually prioritize evaluation criteria. Consequently, this study investigates supervisor-derived criterion weights in thesis assessment and evaluates their impact on AI-based assessment. We surveyed 84 thesis supervisors across four academic disciplines and collected weighting data for 35 thesis assessment criteria. Comparison with the default criterion weights of the AI assessment system RubiSCoT~\cite{Froehlich2026} revealed substantial divergences between supervisor-derived and default criterion weights. To evaluate the practical implications of these differences, the supervisor-derived weights were integrated into multiple calibration configurations and evaluated on a corpus of 80 German-language theses. The best-performing configuration reduced the mean relative deviation between AI-generated and supervisor-assigned evaluations from 11.18\% to 10.85\%, although the improvement was not statistically significant. Human supervisors showed substantially stronger agreement with each other, exhibiting a mean inter-supervisor relative deviation of 4.44\%. The findings indicate that criterion-weight calibration alone does not substantially improve alignment between AI-generated and human assessments.

}


\maketitle

\section{Introduction}
\label{sec:introduction}

Rubrics are widely used in higher education to clearly define how student work is evaluated, support transparency, and improve consistency in assessment \cite{Brookhart2018}. In analytic rubrics, student work is evaluated across multiple criteria rather than through a single holistic judgment, thereby supporting more detailed feedback and more transparent assessment decisions \cite{Brookhart2018}. In thesis assessment, rubric criteria commonly include dimensions such as problem formulation, literature engagement, methodological rigor, argumentation quality, interpretation of results, and reflection on limitations.

Despite the formalization provided by rubrics, thesis assessment remains a complex evaluative process that cannot be fully reduced to written criteria alone \cite{Mullins2002,ODonovan2004}. Supervisors interpret rubric criteria through disciplinary expectations, assessment experience, and implicit academic standards developed through practice \cite{ODonovan2004,Bloxham2012}.

In both human and AI-supported assessment systems, criterion weights determine how strongly individual rubric criteria contribute to assessment outcomes and which evaluation criteria are prioritized during assessment \cite{Shermis2013,Froehlich2026}. However, these weights are typically defined through expert judgment or implementation assumptions rather than empirical evidence regarding actual supervisor weighting behavior. As a result, AI assessment systems may apply rubrics consistently while still diverging systematically from human assessment behavior \cite{Hashemi2024,Froehlich2026}.

Previous research has shown that both human supervisors and AI-supported assessment systems may apply shared rubric criteria differently due to interpretive variability, implicit standards, and calibration limitations \cite{Lundstrom2021,Gingerich2022,Pack2024,Mathews2025,Hashemi2024}. While calibration approaches have been proposed to improve alignment between AI-generated and human evaluations \cite{Hashemi2024}, little research has systematically examined how thesis supervisors prioritize individual rubric criteria when evaluating bachelor's and master's theses. Without empirical evidence regarding supervisor weighting behavior, AI assessment systems risk relying on default criterion weights that may not reflect actual supervisory priorities.

To address this gap, our study presents an empirical analysis of criterion weights reported by thesis supervisors during thesis assessment. Based on a survey of 84 thesis supervisors, we derive criterion weights for rubric criteria distributed across five common thesis components:  \textit{Introduction}, \textit{Literature Review}, \textit{Research Design}, \textit{Results and Discussion}, and \textit{Conclusion}. We then compare these empirically derived criterion weights with the default criterion weights used in the AI assessment system RubiSCoT~\cite{Froehlich2026}.

To evaluate the practical implications of the observed divergences, we integrate the empirically derived criterion weights into the AI assessment system and evaluate multiple calibration configurations on a corpus of 80 German-language theses. For each configuration, we measure the relative deviation between AI-generated and supervisor-assigned assessments.

The contributions of this study are as follows:

\begin{itemize}
\item We provide the first empirical dataset of criterion weights reported by thesis supervisors for thesis assessment across five thesis chapters and four academic disciplines, and release the underlying survey data for reproducibility and future research.\footnote{\url{https://github.com/thorstenfroehlich/RubiSCoT_Validation_Study_Data}}
\item We present a criterion-level analysis of divergences between empirically derived and default AI criterion weights.
\item We evaluate multiple supervisor-derived weighting configurations to analyze their alignment with supervisor-assigned assessments.
\end{itemize}

The remainder of this paper is structured as follows: Section~\ref{sec:related_work} reviews related work on rubric-based thesis assessment, human assessment behavior, and AI-supported assessment systems. Section~\ref{sec:methodology} describes the survey design, corpus construction, and experimental methodology. Section~\ref{sec:results} presents the supervisor-derived criterion weights, human--AI weight divergences, and assessment alignment results across weighting configurations. Section~\ref{sec:discussion} discusses implications, limitations, and interpretation of the findings. Section~\ref{sec:conclusion} concludes the paper and outlines directions for future research.

\section{Related Work}
\label{sec:related_work}

This section reviews prior research on rubric-based thesis assessment, human assessment behavior, and AI-supported assessment systems.

\subsection{Rubric-Based Thesis Assessment}

Rubrics are widely used in higher education to support transparency, consistency, and structured evaluation \cite{Brookhart2018}. In thesis assessment, analytic rubrics commonly evaluate multiple dimensions of academic work, including problem formulation, literature engagement, methodological rigor, argumentation quality, interpretation of results, and reflection on limitations.

Despite the formalization provided by rubrics, thesis assessment remains partly interpretive and shaped by disciplinary expectations, assessment experience, and implicit academic standards \cite{Mullins2002,ODonovan2004,Bloxham2012,Gingerich2022}. Consequently, assessment decisions may still differ substantially despite shared rubric structures.

While prior research has examined rubric reliability, assessment consistency, and rubric design, comparatively little work has investigated how supervisors prioritize individual rubric criteria during thesis assessment. To our knowledge, no previous study has systematically quantified criterion-level weighting behavior across thesis chapters and academic disciplines or examined how empirically derived criterion weights affect alignment between AI-generated and human assessments.

\subsection{Human Assessment Behavior and Criterion Weighting}

Research on assessment behavior has shown that assessors frequently rely on implicit standards and personal benchmarks when evaluating student work \cite{ODonovan2004,Gingerich2022}. Consequently, supervisors may differ substantially in how strongly they prioritize specific criteria despite using the same rubric.

Previous studies have shown that assessors may vary in their evaluation of analytical depth, originality, methodological rigor, and theoretical contribution \cite{Lundstrom2021}. Such differences are particularly relevant in thesis assessment because thesis evaluation involves multidimensional and interpretive judgment. Prior research on AI-supported assessment has similarly reported substantial variability between human evaluators.

Existing research has primarily focused on assessment reliability, rubric interpretation, and evaluator consistency rather than explicitly measuring criterion-level weighting behavior. As a result, little empirical evidence exists regarding how supervisors allocate importance across rubric criteria when assessing bachelor's and master's theses. Understanding these weighting decisions is particularly relevant for AI-supported assessment systems that rely on predefined criterion weights.

\subsection{AI Assessment Systems and Calibration}

AI-supported assessment systems have increasingly been applied to rubric-based evaluation tasks in higher education \cite{Shermis2013,Pack2024,Mathews2025}. Recent advances in large language models have further expanded the capabilities of AI-supported assessment for evaluating complex written assignments, including essays, reports, and theses \cite{Pack2024,Mathews2025}.

Most AI assessment systems use criterion weights to determine how strongly individual evaluation criteria contribute to assessment outcomes. However, these weights are typically defined through expert judgment or implementation assumptions rather than empirical evidence regarding actual supervisor weighting behavior \cite{Hashemi2024,Froehlich2026}.

Recent research has emphasized the importance of calibration approaches that align AI-generated assessments more closely with human judgment \cite{Hashemi2024}. Existing calibration work has primarily focused on prompt engineering, score normalization, and supervised fine-tuning to improve agreement between AI-generated and human evaluations \cite{Hashemi2024,Pack2024}. In contrast, comparatively little attention has been paid to calibration through criterion-weight adjustment.

This represents an important research gap as criterion weights determine how criterion-level evaluations are aggregated within rubric-based assessment systems. Without empirical evidence regarding supervisor weighting behavior, AI assessment systems may rely on weighting schemes that do not reflect actual supervisory priorities. Consequently, even technically consistent AI assessment systems may systematically diverge from human assessment behavior.

\section{Methodology}
\label{sec:methodology}

This section describes the survey design, weighting analysis, evaluation corpus, and calibration configurations used in the study.

\subsection{Study Design}

\begin{figure}[ht]
\centering
\includegraphics[width=0.5\linewidth]{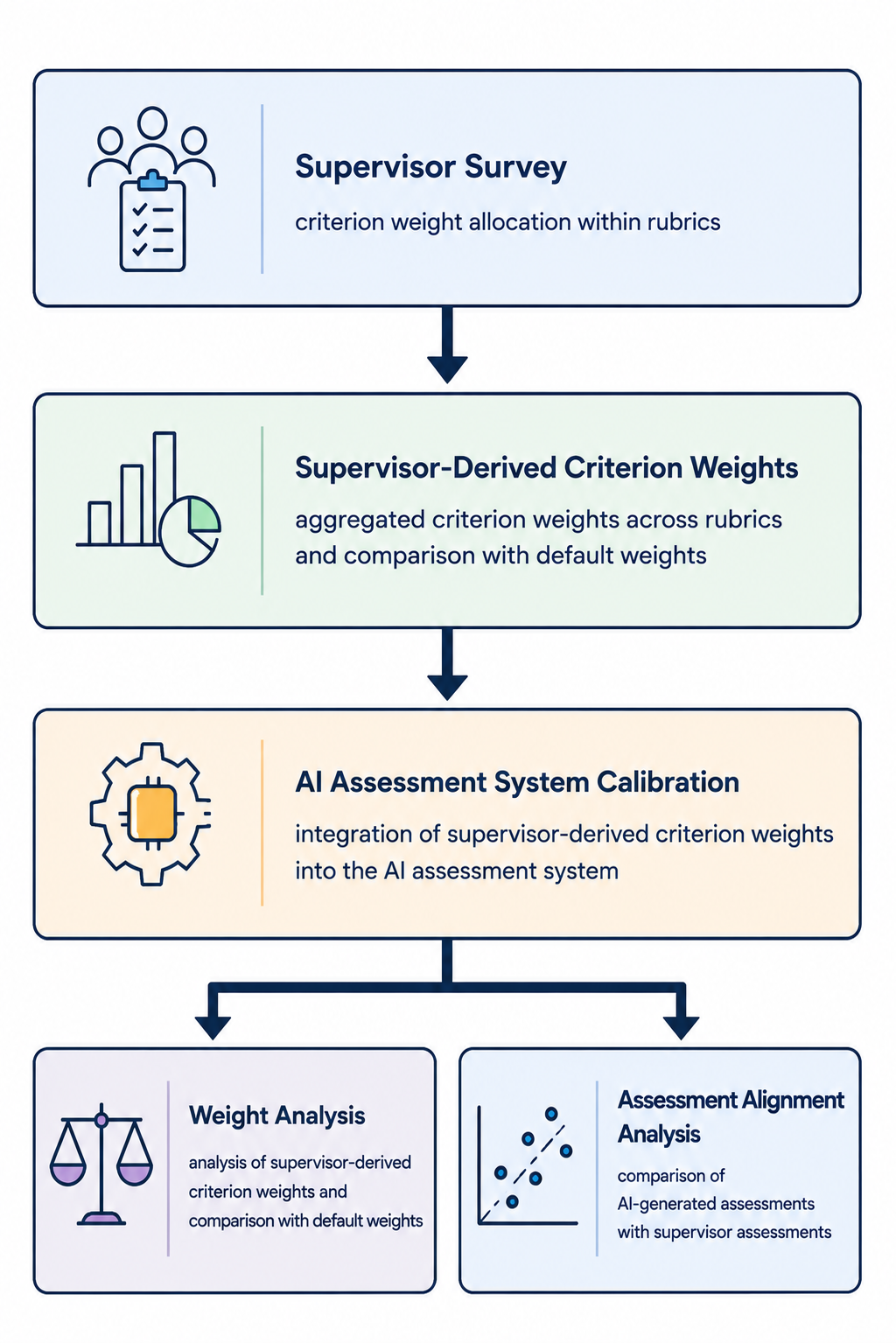}
\caption{Overview of the study design and evaluation workflow.}
\label{fig:study_design}
\end{figure}

Figure \ref{fig:study_design} summarizes the overall study design. We first conducted a supervisor survey in which participants allocated criterion weights within thesis assessment rubrics. Based on these responses, we derived aggregated criterion weights across rubric categories.

The study then proceeded along two complementary evaluation paths. First, we conducted a weight analysis by comparing the criterion weights reported by thesis supervisors with the default criterion weights used in the AI assessment system. This analysis quantified where human assessment priorities and default AI criterion weights diverged.

Second, we integrated the empirically derived criterion weights into the AI assessment system and evaluated whether recalibrated weighting configurations produced chapter-level scores that were closer to supervisor-assigned assessments. For this purpose, we executed multiple calibration configurations on a corpus of thesis submissions and compared the resulting AI-generated scores with the corresponding supervisor assessments.

\subsection{Supervisor Survey}

We conducted an online survey to collect criterion weights reported by thesis supervisors for thesis assessment rubrics. The survey was implemented using Qualtrics and replicated the rubric structure used in the AI assessment system.

The rubric consisted of 35 evaluation criteria distributed across five core thesis components: \textit{Introduction}, \textit{Literature Review}, \textit{Research Design}, \textit{Results and Discussion}, and \textit{Conclusion}. These components represent common functional elements of academic theses rather than fixed chapter titles. While individual theses may use different headings or organize content differently, the corresponding content can generally be mapped to these five areas during assessment. Throughout the remainder of this paper, these components are referred to as \textit{thesis chapters} for simplicity.

To measure relative criterion importance, we used a constant-sum allocation design \cite{Skedgel2015}. Participants distributed 100\% across the criteria within each chapter rubric, forcing explicit prioritization decisions and avoiding ceiling effects commonly observed in Likert-scale importance ratings. All criterion labels and descriptions matched those used in the AI assessment system to ensure direct comparability. Using the original rubric wording also minimized wording differences between the survey instrument and the AI assessment system.

After each weighting section, participants could optionally provide qualitative explanations for their weighting decisions. Demographic information included discipline, supervision experience, grading language, degree levels supervised, and AI tool usage. Participants were recruited through academic mailing lists, professional networks, and social media channels between March 13 and March 25, 2026. 
The final sample of 84 thesis supervisors included supervisors from Business and Economics ($n=32$), Social Sciences ($n=25$), STEM disciplines ($n=19$), and Humanities ($n=6$). Two additional participants selected ``Other'' as their primary discipline. Most participants reported extensive thesis supervision experience and primarily assessed German-language theses.

RubiSCoT is an LLM-based thesis assessment system that evaluates theses using a structured analytic rubric. The system assesses thesis components separately, generates criterion-level scores and explanations, and aggregates these scores into chapter-level assessments using predefined criterion weights. In this study, RubiSCoT serves as the AI assessment system whose default criterion weights are compared with supervisor-derived criterion weights.

\begin{figure*}[t]
    \centering
    \includegraphics[width=0.95\textwidth]{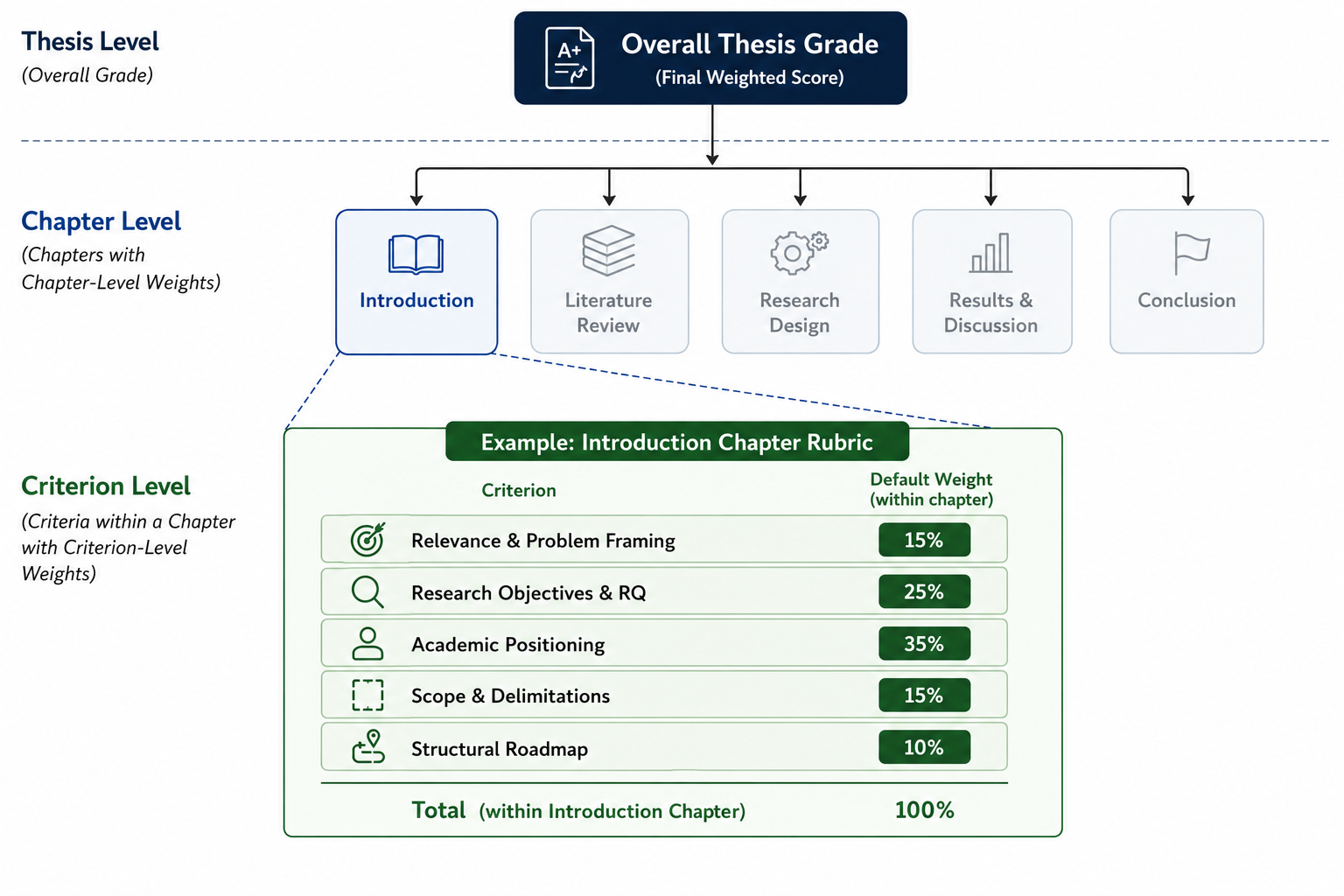}
    \caption{Hierarchical structure of RubiSCoT’s rubric weighting system. Shown weights are for illustration only.}
    \label{fig:rubric_structure}
\end{figure*}

Figure~\ref{fig:rubric_structure} illustrates RubiSCoT’s hierarchical rubric structure. While the system contains both chapter-level and criterion-level weights, the present study focuses exclusively on criterion-level weights since chapter-level weights are often institutionally predefined and cannot easily be modified within operational assessment frameworks. The figure shows the default criterion-weight distribution for the \textit{Introduction} chapter as an example.

The hierarchical weighting structure forms the basis for the calibration configurations evaluated in this study. The default criterion weights used in the AI assessment system were originally defined during the development of RubiSCoT based on expert judgment and iterative rubric design decisions \cite{Froehlich2026}. These weights originate from the production version of RubiSCoT and are not reported in the original system description. The complete default criterion-weight vector for all five rubrics is available in our accompanying repository (see Section~\ref{sec:introduction}). They reflect the assumed relative importance of rubric criteria during thesis assessment but had not previously been validated against empirical supervisor weighting data.

\subsection{Weight Analysis}

We aggregated the supervisor responses to derive mean criterion weights for all rubric criteria.
For each criterion, we calculated the mean weight, standard deviation, and coefficient of variation across all supervisors.
These statistics describe both the average importance assigned to each criterion and the degree of agreement among supervisors.
Unlike standard deviation alone, the coefficient of variation normalizes variability relative to the mean weight, thereby enabling direct comparison across criteria with different average weights.
Lower values indicate stronger agreement among supervisors.

To quantify differences between supervisor-derived criterion weights and the default criterion weights used in the AI assessment system, we calculated the mean absolute error (MAE):

\begin{equation}
\text{MAE (\%)} =
\frac{1}{n}
\sum_{i=1}^{n}
\abs{\text{w}_{\text{supervisor},\text{i}} - \text{w}_{\text{default},\text{i}}}
\times 100
\end{equation}

where $n$ denotes the number of criteria within a rubric,
$\text{w}_{\text{supervisor},\text{i}}$ represents the aggregated supervisor-derived criterion weight,
and $\text{w}_{\text{default},\text{i}}$ represents the default criterion weight used in the AI assessment system.

Lower MAE values indicate stronger alignment between supervisor-derived and default criterion weights.

We used MAE rather than relative percentage deviation because criterion weights sum to 100\% within each rubric. Relative deviations can be misleading for small weights, where modest absolute differences may produce large percentage changes. MAE therefore provides a more stable and interpretable measure of divergence between two criterion-weight distributions.

This analysis enabled a criterion-level comparison between human assessment priorities and the default criterion-weight distribution of the AI assessment system.

\subsection{Evaluation Corpus}

The evaluation corpus consists of 80 German-language theses from IU International University of Applied Sciences. The corpus includes 20 BA, 20 B.Sc., 20 MA, and 20 M.Sc. theses and covers both technical and non-technical disciplines. 
The theses varied substantially in length, ranging from 47 to 335 pages (average: 110.1 pages; median: 99.5 pages).

Each thesis was previously assessed by one or two human supervisors whose evaluations serve as the reference for comparison. 
For each thesis, chapter-level reference scores were available from the rubric-based assessment forms used during thesis evaluation. These chapter-level scores served as the human reference evaluations for all alignment analyses. The survey was anonymous and distributed through public channels, whereas the corpus was graded by supervisors of IU International University of Applied Sciences during routine thesis assessment without reference to RubiSCoT or its criterion weights. Both data sources therefore reflect thesis assessment practice, but they do not model the preferences of identical individual supervisors.
For the subset of 70 theses with two available supervisor evaluations, we additionally calculated inter-supervisor chapter-level deviations to establish a human assessment baseline for comparison with AI-generated assessments. Across all chapter evaluations, the mean relative deviation between first and second supervisors was 4.44\%.

\subsection{Calibration Configurations}

We evaluated five weighting configurations.

\subsubsection{Default Configuration}
The original default criterion weights of the AI assessment system were applied to the complete thesis corpus.

\subsubsection{Global Calibration}
We calculated one shared supervisor-derived criterion-weight distribution using all survey responses.
This configuration was applied to the complete thesis corpus.

\subsubsection{Degree-Level Calibration}
We calculated separate criterion-weight distributions for bachelor-level and master-level supervision.
The bachelor-level configuration was applied only to BA and B.Sc. theses, while the master-level configuration was applied only to MA and M.Sc. theses.

\subsubsection{Discipline-Level Calibration}
We calculated separate criterion-weight distributions for technical and non-technical disciplines.
The technical configuration was applied only to B.Sc. and M.Sc. theses, while the non-technical configuration was applied only to BA and MA theses.

\subsubsection{Combined Calibration}
We calculated four specialized criterion-weight distributions:
BA,
B.Sc.,
MA,
and M.Sc.
Each configuration was applied exclusively to the corresponding thesis group.

\subsection{Chapter-Level Alignment Evaluation}

For each calibration configuration, we executed the AI assessment system on the corresponding subset of theses. The system generated chapter-level assessment scores using the respective criterion-weight distribution.

To isolate the effect of criterion-weight calibration, we focused the primary evaluation on chapter-level score alignment rather than final thesis grades. Chapter-level scores provide a more direct measure of the impact of criterion weights, whereas final thesis grades additionally depend on institution-specific chapter weights and grading systems.

We therefore compared the AI-generated chapter-level scores with the corresponding supervisor-assigned chapter-level scores across the five thesis chapters.
Evaluation focused on the mean relative deviation between AI-generated and supervisor-assigned chapter scores:

\begin{equation}
\text{Relative Deviation (\%)} =
\frac{\abs{\text{s}_{\text{AI}} - \text{s}_{\text{human}}}}{\text{s}_{\text{human}}} \times 100
\end{equation}

where s$_{\text{AI}}$ denotes the AI-generated chapter-level score and s$_{\text{human}}$ denotes the corresponding supervisor-assigned chapter-level score. Lower deviation indicates stronger alignment with human scoring behavior.

To evaluate whether the improvement of the best-performing calibration configuration over the default configuration was statistically significant, we conducted significance testing on thesis-level aggregated chapter deviations. Since the same theses were evaluated across configurations, all tests used paired observations. We additionally calculated Cohen’s $d$ to estimate effect sizes. Statistical significance was evaluated at the $\alpha = 0.05$ level.

To avoid treating chapter evaluations from the same thesis as statistically independent observations, statistical testing was conducted on thesis-level aggregated deviations rather than on individual chapter-level deviations. For each thesis, the mean relative chapter-level deviation across all five chapters was calculated separately for each calibration configuration.

\section{Results}
\label{sec:results}

This section presents the supervisor-derived criterion weights, human-AI weighting divergences, and chapter-level alignment results across calibration configurations.

\subsection{Human-AI Weight Divergence}

We compared the criterion weights reported by thesis supervisors with the default criterion weights used in the AI assessment system. 
Table \ref{tab:mae_by_chapter} summarizes the mean absolute error (MAE) between both weighting schemes across the five thesis rubrics. The reported MAE values represent the average absolute difference between supervisor-derived and default criterion weights within each rubric. Lower MAE values therefore indicate stronger agreement between human and AI weighting priorities.

\begin{table}[ht]
\centering
\caption{Mean Absolute Error Between Supervisor-Derived and Default Criterion Weights}
\label{tab:mae_by_chapter}

\setlength{\tabcolsep}{2.3pt}

\footnotesize

\begin{tabular}{@{}lcl@{}}
\toprule
\textbf{Chapter} 
& \textbf{Mean Absolute Error (\%)} 
& \textbf{Largest Criterion Difference (\%)} \\
\midrule

\textit{Introduction}
& 6.92
& \textit{Academic Positioning} ($-16.0$) \\

\textit{Literature Review}
& 6.72
& \textit{Analytical Synthesis} ($-15.5$) \\

\textit{Research Design}
& 1.70
& \textit{Reflexivity} ($+4.8$) \\

\textit{Results and Discussion}
& 2.58
& \textit{Derivation} ($-4.9$) \\

\textit{Conclusion}
& 3.59
& \textit{Synthesis} ($-8.4$) \\

\midrule

\textbf{Overall}
& \textbf{4.30}
& \\

\bottomrule
\end{tabular}
\end{table}

The largest divergences occurred in the \textit{Introduction} and \textit{Literature Review} rubrics, with MAE values of 6.92\% and 6.72\%, respectively. In both cases, the default AI assessment system emphasized higher-order analytical criteria more strongly than supervisors did.

Within the \textit{Introduction} rubric, the default AI assessment system allocated substantially more weight to the \textit{Academic Positioning} criterion ($-16.0$\%) than supervisors. Supervisors instead assigned greater importance to problem framing and relevance of literature sources.

Similarly, within the \textit{Literature Review} rubric, the AI assessment system emphasized \textit{Analytical Synthesis} ($-15.5$\%) more strongly than supervisors, who instead prioritized relevance and coverage of literature sources.

In contrast, the \textit{Research Design} (MAE = 1.70\%) and \textit{Results and Discussion} (MAE = 2.58\%) rubrics showed comparatively strong alignment, emphasizing methodological quality, analytical rigor, and evidence-based interpretation of results.

Overall, the findings indicate that the largest human--AI weighting divergences occurred in more interpretive evaluation criteria, whereas methodological and technical criteria showed substantially stronger alignment.

\subsection{Chapter-Level Alignment Results}

We evaluated whether supervisor-derived criterion weights improved the alignment between AI-generated and supervisor-assigned chapter-level assessments across the five thesis chapters.

Table~\ref{tab:configuration_overview} summarizes the evaluated calibration configurations, while Table~\ref{tab:chapter_alignment} reports the corresponding mean relative chapter-level deviations.

\begin{table}[ht]
\centering
\caption{Overview of the Evaluated Calibration Configurations}
\label{tab:configuration_overview}

\setlength{\tabcolsep}{2.3pt}

\footnotesize

\begin{tabular}{@{}lll@{}}
\toprule
\textbf{Configuration} 
& \textbf{Supervisor-derived Weights} 
& \textbf{Evaluated Theses} \\
\midrule

\textit{Default}
& Default AI system weights
& Full Corpus \\

\midrule

\textit{Global}
& All supervisors
& Full Corpus \\

\midrule

\textit{Degree-Level}
& Bachelor supervisors
& BA + B.Sc. \\

& Master supervisors
& MA + M.Sc. \\

\midrule

\textit{Discipline-Level}
& Technical supervisors
& B.Sc. + M.Sc. \\

& Non-technical supervisors
& BA + MA \\

\midrule

\textit{Combined}
& Non-technical BA supervisors
& BA \\

& Technical B.Sc. supervisors
& B.Sc. \\

& Non-technical MA supervisors
& MA \\

& Technical M.Sc. supervisors
& M.Sc. \\

\bottomrule
\end{tabular}
\end{table}

\begin{table}[ht]
\centering
\caption{Chapter-Level Deviations Between AI-Generated and Supervisor-Assigned Scores}
\label{tab:chapter_alignment}

\setlength{\tabcolsep}{1.2pt}
\footnotesize

\begin{tabular}{@{}lc|cccc|c@{}}
\toprule
\textbf{Chapter}
& \textit{Default~}
& \textit{~Global}
& \textit{Degree-Level}
& \textit{Discipline-Level}
& \textit{Combined~}
& \textit{~Supervisors} \\
\midrule

\textit{Introduction}
& 9.94
& 9.76
& 9.90
& 9.55
& 9.91
& \textbf{2.51} \\

\textit{Literature Review}
& 11.96
& 12.36
& 12.78
& 12.30
& 12.14
& \textbf{3.78} \\

\textit{Research Design}
& 11.50
& 11.50
& 11.17
& 11.95
& 11.20
& \textbf{5.23} \\

\textit{Results and Discussion}
& 10.91
& 10.67
& 10.35
& 10.37
& 9.81
& \textbf{4.84} \\

\textit{Conclusion}
& 11.61
& 11.52
& 12.18
& 11.16
& 11.17
& \textbf{5.85} \\

\midrule

\textbf{Overall}
& 11.18
& 11.16
& 11.28
& 11.07
& \textbf{10.85}
& \textbf{4.44} \\

\bottomrule
\end{tabular}
\end{table}

\subsubsection{Results with Global Calibration}

The \textit{Global} configuration applied one shared supervisor-derived criterion-weight distribution across the complete thesis corpus.

Compared to the \textit{Default} configuration (11.18\%), the \textit{Global} configuration achieved a slightly lower overall mean relative chapter-level deviation of 11.16\%. Improvements were primarily observed for the \textit{Introduction}, \textit{Results and Discussion}, and \textit{Conclusion} chapters, while deviations in \textit{Literature Review} increased slightly.

\subsubsection{Results with Degree-Level Calibration}

The \textit{Degree-Level} configuration used separate criterion-weight distributions for bachelor-level and master-level theses. The bachelor-level configuration was applied to BA and B.Sc. theses, while the master-level configuration was applied to MA and M.Sc. theses.

The \textit{Degree-Level} configuration produced an overall mean relative chapter-level deviation of 11.28\%, slightly higher than the \textit{Default} configuration. Improvements in \textit{Research Design} and \textit{Results and Discussion} were offset by larger deviations in \textit{Literature Review} and \textit{Conclusion}.

\subsubsection{Results with Discipline-Level Calibration}

The \textit{Discipline-Level} configuration used separate criterion-weight distributions for technical and non-technical disciplines. The technical configuration was applied to B.Sc. and M.Sc. theses, while the non-technical configuration was applied to BA and MA theses.

The \textit{Discipline-Level} configuration achieved an overall mean relative chapter-level deviation of 11.07\%, representing the second-best result among all evaluated configurations. The strongest improvements were observed for \textit{Introduction}, \textit{Results and Discussion}, and \textit{Conclusion}, whereas \textit{Research Design} showed slightly larger deviations than under the \textit{Default} configuration.

\subsubsection{Results with Combined Calibration}

The \textit{Combined} configuration used separate criterion-weight distributions for BA, B.Sc., MA, and M.Sc. theses. Each configuration was applied exclusively to the corresponding thesis group.

The \textit{Combined} configuration achieved the strongest overall alignment, reducing the mean relative chapter-level deviation from 11.18\% to 10.85\%. This corresponds to a relative improvement of 2.95\%.

The strongest improvement was observed for \textit{Results and Discussion}, where the relative deviation decreased from 10.91\% to 9.81\%, corresponding to a relative improvement of 10.1\%. Additional improvements were observed for \textit{Research Design} and \textit{Conclusion}.

\subsubsection{Overall Configuration Comparison}

Across all evaluated configurations, the \textit{Combined} calibration achieved the lowest overall chapter-level deviation (10.85\%), followed by the \textit{Discipline-Level} (11.07\%) and \textit{Global} (11.16\%) configurations. The \textit{Degree-Level} configuration produced the largest overall deviation (11.28\%).

Despite these improvements, all AI configurations remained substantially less aligned with supervisor assessments than human supervisors were with each other. The mean inter-supervisor relative deviation was 4.44\%, compared to 10.85\% for the best-performing AI configuration.

Our paired t-test ($\alpha=.05$) comparing the \textit{Default} and \textit{Combined} configurations found no statistically significant difference in mean relative chapter-level deviation ($t(79)=0.75$, $p=.458$, $d=0.08$).

Although the \textit{Combined} configuration achieved the lowest overall deviation, the observed reduction remained within the range of sampling variability (95\% CI [-0.57\%, 1.26\%]). These results indicate that the observed improvement is small relative to the variability across theses and therefore cannot be distinguished from random variation at the $\alpha=.05$ significance level.

\section{Discussion}
\label{sec:discussion}

This section interprets the findings and discusses their implications for AI-supported thesis assessment.

\subsection{Human-AI Weight Divergence}

The comparison between supervisor-derived and default AI criterion weights revealed a clear pattern. Divergences were largest in the more interpretive \textit{Introduction} and \textit{Literature Review} rubrics, whereas stronger alignment was observed in \textit{Research Design}. 
This pattern suggests that human supervisors and AI assessment systems differ most strongly in areas requiring interpretive judgment, whereas methodological dimensions appear easier to operationalize consistently.

The findings further indicate that the default AI weighting scheme places greater emphasis on higher-order analytical criteria, whereas supervisors assigned greater importance to problem framing, source relevance, and methodological quality. This suggests that default AI weighting schemes may reflect assessment priorities that differ from those of human supervisors.

\subsection{Impact of Empirical Weight Alignment}

The results show that empirically derived criterion weights did not produce a detectable improvement between AI-generated and supervisor-assigned assessments. The \textit{Combined} configuration performed best, reducing the mean relative chapter-level deviation from 11.18\% to 10.85\%.

However, the improvement was small, not statistically significant, and substantially lower than the agreement observed between human supervisors. The mean inter-supervisor relative deviation was 4.44\%, compared to 10.85\% for the \textit{Combined} configuration.

These findings suggest that reducing criterion-weight divergence alone is insufficient to substantially improve alignment between AI-generated and human assessments.

One possible explanation is that differences arise less from criterion weights than from how human supervisors and AI systems interpret criteria, evaluate evidence, and form assessment judgments.

The strongest improvements were observed in the \textit{Results and Discussion} chapter. This may indicate that weighting adjustments are most effective in chapters where evaluation criteria are well defined and closely connected to explicit analytical tasks. In contrast, the stable deviations observed in \textit{Literature Review} suggest that factors beyond criterion weighting play a larger role in chapters requiring broader interpretive judgment.

\subsection{Implications for AI-Supported Assessment Systems}

The findings have two implications for AI-supported assessment systems:

First, the observed divergences indicate that default criterion weights should not be assumed to reflect actual supervisory priorities without empirical validation.

Second, the limited impact of weight calibration suggests that AI-supported assessment systems should not be calibrated through criterion weights alone. Future calibration should also address rubric interpretation, evidence evaluation, and assessment reasoning, for example through improved prompts, reasoning traces, or human-in-the-loop calibration.

\subsection{Limitations}

Several limitations should be considered when interpreting the findings:

First, the study relies on self-reported supervisor weights rather than directly observed assessment behavior. Although the constant-sum allocation design forced supervisors to make explicit trade-offs between criteria, actual assessment decisions may involve more dynamic weighting processes. In addition, self-reported weights may be influenced by social desirability bias or normative assumptions regarding good assessment practice.

Second, the supervisor sample was geographically concentrated in Germany and primarily involved German-language thesis assessment. As assessment cultures and thesis expectations vary across national and institutional contexts, the findings may not fully generalize to other higher education systems.

Third, some subgroup analyses were based on relatively small samples, particularly for specialized combinations in the \textit{Combined} configuration, which may reduce the stability of highly specialized criterion-weight distributions.

Fourth, the study evaluates alignment at the chapter level rather than individual rubric criteria. While chapter-level assessments provide a meaningful basis for comparing human and AI evaluations, criterion-level assessment data from human supervisors were not available for the thesis corpus.

Finally, the study focused exclusively on criterion weighting and did not evaluate other components of AI-supported assessment systems, such as rubric interpretation, textual reasoning quality, or cross-chapter consistency analysis.

\section{Conclusion and Future Work}
\label{sec:conclusion}

This section summarizes the findings and discusses possible future work.

\subsection{Conclusion}

This study presented an empirical investigation of criterion weighting in thesis assessment. Using a survey of 84 thesis supervisors across four academic disciplines, we collected criterion weights for 35 rubric criteria distributed across the five core thesis components: \textit{Introduction}, \textit{Literature Review}, \textit{Research Design}, \textit{Results and Discussion}, and \textit{Conclusion}.

The findings revealed substantial divergences between supervisor-derived and default AI criterion weights, particularly in the Introduction and Literature Review rubrics, whereas methodological criteria showed substantially stronger alignment.

To evaluate the practical implications of these divergences, we integrated supervisor-derived criterion weights into the AI assessment system and evaluated multiple weighting configurations on 80 German-language theses.

The best-performing configuration reduced the mean relative chapter-level deviation from 11.18\% to 10.85\%. However, the improvement was not statistically significant, and human supervisors showed substantially stronger agreement with each other (4.44\%) than with AI-generated assessments. Overall, the findings indicate that criterion-weight calibration alone is insufficient to substantially improve human--AI alignment in thesis assessment.

This study contributes an empirical dataset of supervisor-derived criterion weights, a reproducible methodology for human--AI weighting comparison, and an experimental framework for evaluating assessment alignment in AI-supported thesis assessment systems.

\subsection{Future Work}

Several directions for future research emerge from these findings.

First, future studies could investigate whether criterion-weight distributions differ across countries, institutions, or disciplinary cultures, and whether AI-supported assessment systems require context-specific weighting strategies.

Second, future research should investigate additional sources of human--AI divergence, including differences in rubric interpretation, evidence evaluation, and assessment reasoning.

Third, future research should investigate whether limited variation in criterion-level scores constrains the effect of criterion-weight calibration on chapter-level scores.

Fourth, future work could explore dynamic weighting approaches in which criterion importance adapts to thesis characteristics or intermediate evaluation signals rather than relying on fixed weighting configurations.

Finally, future research should investigate how calibration strategies influence fairness, transparency, and trust in AI-supported assessment systems. Similar calibration challenges may also emerge in other AI-supported assessment domains, including automatic short-answer assessment \cite{Schlippe2021}.

\section*{Acknowledgments}

We thank the 84 thesis supervisors who contributed their expertise by participating in the survey.
This research was conducted as part of the FAIRGRADE project at the Research Institute Artificial Intelligence (AI) of IU International University of Applied Sciences.

\bibliography{00_main}% common bib file


\end{document}